%% file: root.tex
\documentclass[letterpaper, 10 pt, conference]{ieeeconf}  

\IEEEoverridecommandlockouts                              

\usepackage{graphics} 
\usepackage{epsfig} 
\usepackage{times} 
\usepackage{amsmath} 
\usepackage{amssymb}  
\usepackage{amsfonts}

\let\emptyset\varnothing
\title{\LARGE \bf
DFineNet: Ego-Motion Estimation and Depth Refinement from Sparse, Noisy Depth Input with RGB Guidance 
}

\author{Yilun Zhang, Ty Nguyen, Ian D. Miller, Shreyas S. Shivakumar, \\
Steven Chen, Camillo J. Taylor, Vijay Kumar
\thanks{This work was supported in part by the Semiconductor Research Corporation (SRC)
and DARPA.}
\thanks{The authors are with the
GRASP Lab, University of Pennsylvania, Philadelphia, PA 19104 USA. {
{\tt\footnotesize email: \{zhyilun, tynguyen, iandm, sshreyas, chenste, cjtaylor, kumar\}}@seas.upenn.edu}}%
}

\begin{document}

\maketitle

\thispagestyle{empty}
\pagestyle{empty}

\begin{abstract}
Depth estimation is an important capability for autonomous vehicles to understand and reconstruct 3D environments as well as avoid obstacles during the execution. Accurate depth sensors such as LiDARs are often heavy, expensive and can only provide sparse depth while lighter depth sensors such as stereo cameras are noiser in comparison. We propose an end-to-end learning algorithm that is capable of using sparse, noisy input depth for refinement and depth completion. Our model also produces the camera pose as a byproduct, making it a great solution for autonomous systems. We evaluate our approach on both indoor and outdoor datasets. Empirical results show that our method performs well on the KITTI~\cite{kitti_geiger2012we} dataset when compared to other competing methods, while having superior performance in dealing with sparse, noisy input depth on the TUM~\cite{sturm12iros} dataset. \end{abstract}

\input{tex/1_Introduction.tex}

\input{tex/2_Related_work.tex}

\input{tex/3_Problem_formulation.tex}

\input{tex/4_Framework.tex}
\input{tex/5_Experiment.tex}

\input{tex/6_Evalutation.tex}

\input{tex/7_Conclusion.tex}

\bibliographystyle{bib/IEEEtran}
\bibliography{bib/bib.bib}

\end{document}

%% file: tex/1_Introduction.tex
\section{Introduction}
For autonomous systems such as Micro Aerial Vehicles (MAVs), depth and ego-motion (pose) estimation are important capabilities. Depth measurements are vital while reasoning in 3D space, which is necessary for environmental understanding and obstacle avoidance, particularly for moving platforms. Ego-motion estimation enables the robot to track its motion, allowing it to follow particular trajectories. However, performing depth estimation on-board MAVs is a challenging task given the size, weight, and power (SWaP) constraints, resulting in additional computational constraints. While accurate depth sensors such as LiDARs are relatively heavy, they provide accurate depth estimates and are often used on larger UAV platforms \cite{fla2017} and are also popular in the autonomous vehicle domain. These depth sensors are often heavy and expensive, limiting their use in MAV applications. In contrast, inexpensive depth sensors such as active and passive stereo cameras are lightweight, but their reduced weight comes at the cost of accuracy. Time-of-Flight (ToF) cameras are viable alternatives to LiDARs in certain situations and have been used on MAVs as they meet the size and weight constraints, but they are often limited in resolution and field of view when compared to LiDARs \cite{shivakumar2018real}.

\label{sec:introduction}
\begin{figure}[t]
    \centering
    \includegraphics[width=0.99\linewidth]{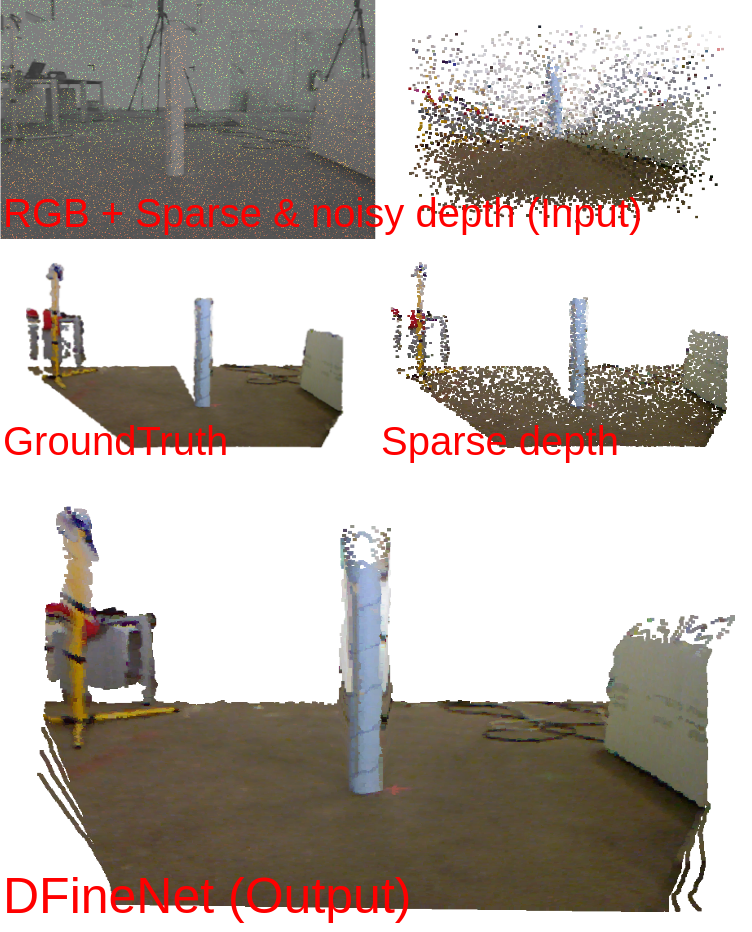}
    \caption{An example of sparse, noisy depth input (1st row), the 3D visualization of ground truth of depth (2nd row) and the 3D visualization of output from our model (bottom). RGB image (1st) is overlaid with sparse, noisy depth input for visualization }
    \label{fig:demo}
\end{figure}

In this study, we propose a method where we use noisy depth measurements from popular low-cost stereo cameras, RGBD sensors, and sparse depth from LiDAR. We present a framework capable of filtering input noise and estimating pose simultaneously, and argue for jointly estimating pose and depth to improve estimations in both domains.

State-of-the-art monocular algorithms such as DSO \cite{engel2016dso} show impressive performance using monocular images only.  However, these methods do not provide an easy way to incorporate depth measurements from other sources.  DSO has be extended to exploit stereo images \cite{wang2017dsostereo}, and other stereo odometry algorithms have been developed such as \cite{wenxin2018}, but these methods do not generalize well to handle wide varieties of depth data, from sparse, precise, LiDAR depth to dense, noisy, ToF depth.

Given the advances in deep neural networks and in particular convolutional neural networks (CNNs), there has been increasing interest in applying CNNs to the depth and ego-motion estimation problem. Unsupervised algorithms such as~\cite{zhou2017unsupervised, undeepvo_li2018, sfmlearner++_prasad2018} are appealing as they utilize only regular camera images to train and are capable of predicting both depth and ego-motion. However, they can only provide depth up to scale.  In addition, the performance of state-of-the-art methods including~\cite{top_depth_prediction_fu2018deep}, which ranks number one in the KITTI depth prediction benchmark at the time of writing, is still far from usable in autonomous robots where accurate depth estimation is crucial for fast, safe execution. 
Recent success in fusing sparse depth measurements and RGB images~\cite{top_depth_completion_van2019sparse,fang_net_2018self} have encouraged the application of depth completion to robotics. However, these methods rely heavily on the sparse depth input and regions where sparse depth information exists, rendering them unable to extrapolate depth in regions where there is no LiDAR measurement. They also isolate the pose and depth estimation problems from each other, which we believe amounts to throwing out information that can be used to improve both estimators. The method proposed by Ma et al.~\cite{fang_net_2018self} uses Perspective-n-Pose (PnP) to estimate pose after first estimating depth, but becomes susceptible to failure in situations with low image texture, given the limitations of PnP. Additionally, the authors assume that the sparse depth input is noiseless, which limits the method's usability with inexpensive noisy depth sensors.  Their method was additionally trained and tested on accurate LiDAR depth information only.

\begin{figure*}[t]
  \includegraphics[width=\textwidth]{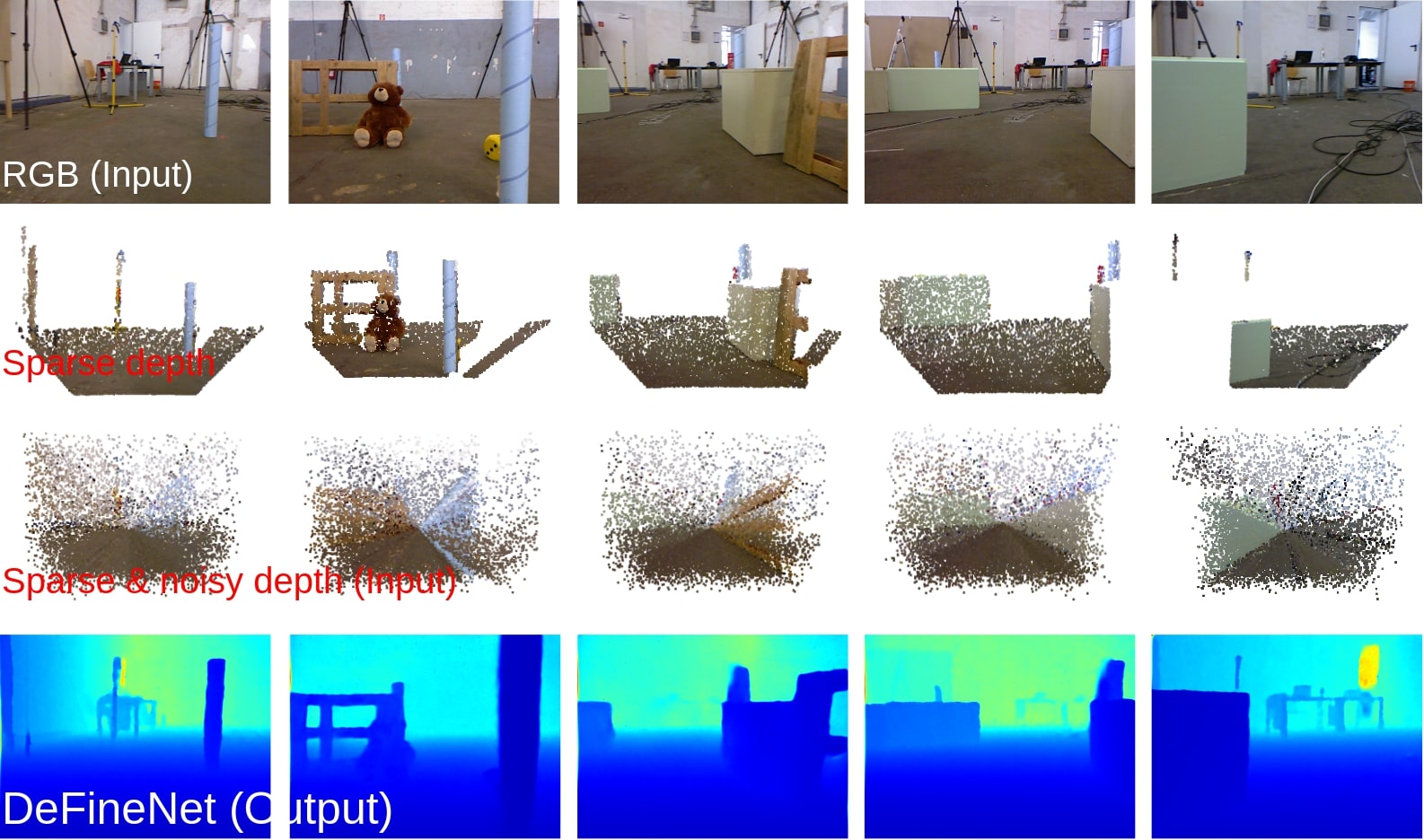}
  \caption{Our method can refine sparse \& noisy depth input (the 3rd row) to output dense depth of high quality (bottom row). }
  \label{fig: showhead}
\end{figure*}

In this work, we argue that the depth prediction CNN should be jointly trained with a pose prediction CNN, resulting in an end-to-end algorithm to fuse the sparse depth and the RGB image. 
Specifically, our contributions are as follows:
\begin{itemize}
    \item We propose an end-to-end framework that can learn to complete and refine depth images (as shown in Fig.~\ref{fig: showhead}) from a variety of depth sensors with high degrees of noise.
    \item Our end-to-end framework learns to simultaneously generate refined, completed depth estimates along with accurate pose estimates of the camera.
    \item We evaluate our proposed approach in comparison with state-of-the-art approaches on indoor and outdoor datasets, using both sparse LiDAR depth as well as dense depth from stereo. 
    \item Experimental results show that our method outperforms other competing methods on pose estimation and depth prediction in challenging environments. 
\end{itemize}

%% file: tex/2_Related_work.tex
\section{Related Work}
\label{sec:related_work}
\subsection{Depth Sensing}
Depth measurements can be obtained from a variety of hardware, including sensors such as LiDARs, Time of Flight (ToF) cameras, and active and passive stereo cameras. With these sensors however, there is typically a trade-off between size, weight and cost on one axis, and accuracy, resolution, and range on the other. Sensors that are accurate, such as 3D LiDARs, are expensive and heavy, while small lightweight sensors such as stereo cameras or structured light sensors have limited range and are often noisier.  

An affordable and popular option among these depth sensors is the stereo camera which, in both active and passive cases, consists of two cameras separated by a baseline distance.  The estimation of depth comes from the measurement of disparity of a 3D point in the two 2D images. Popular stereo reconstruction algorithms such as SGM~\cite{hirschmuller2008stereo} and local matching~\cite{einecke2015multi} perform a triangulation of 3D points from corresponding 2D image points observed by two cameras to obtain the depth of these 3D points. Compared to LiDARs and ToF cameras, stereo cameras can provide denser depth since they are usually only limited by the raw image resolution of the stereo pair. However, their effective range is limited by the baseline, which is inevitably limited by the robot's size. Even the best stereo methods are prone to noise and are susceptible to all the challenges of regular image based perception algorithms. Some of the leading stereo algorithms are compared on the KITTI Stereo 2015~\cite{kitti_geiger2012we} and Middlebury benchmarks~\cite{scharstein2002taxonomy}. 

\subsection{Visual Odometry (Joint Depth and Pose Estimation)}
3D reconstruction can also be performed by triangulating 3D points observed by the same camera at two consecutive time steps. Monocular depth estimation approaches estimate the pose or transformation between two consecutive camera frames either from other sensors such as a GPS or IMU or by a pose estimation process. 

One prominent approach for monocular depth estimation is feature-based visual odometry~\cite{torr1999feature, nister2004visual, badino2013visual} which determines pose from a set of well established correspondences. These methods keep track of only a small set of features, often losing relevant contextual information about the scene. They also suffer from degraded imaging conditions such as image blur and low light where detecting and establishing features is challenging. 

Direct visual odometry methods~\cite{lucas1981iterative, irani1999direct, alismail2017direct, engel2016dso}, on the other hand, directly use pixel intensity values and exploit the locality assumption of the motion to formulate an energy function. The purpose of this energy function is to minimize the reprojection error and thereby obtain the pose of the camera. These approaches generally yield better results then feature-based algorithms, but are more sensitive to changes in illumination between successive frames. Direct visual odometry can also provide a dense depth and is intuitively similar to our approach.  However, purely monocular based approaches suffer from an inherent scale ambiguity.

State-of-the-art approaches to deep-learning based monocular dense depth estimation rely on DCNNs~\cite{cao2018estimating, fu2017compromise}. However, as Godard et al.~\cite{undeepVO_godard2017unsupervised} note, relying solely on image reconstruction for monocular depth estimation results in low quality depth images. However, they mitigate this problem by leveraging spatial constraints between the two images captured by the cameras during training. This method then requires stereo images to be available during training process. 

The problem of scale ambiguity can be solved by utilitzing additional information.  Wang et al.~\cite{wang2017dsostereo} modify DSO \cite{engel2016dso} to take as input a stereo pair of images instead of purely monocular data.  They demonstrate significantly improved depth estimation and odometry, but rely on stereo images and cannot generalize to handle other depth sensors.  Yang et al.~\cite{yang2018deep} improve upon this by incorporating a deep network for depth estimation using not only a left-right consistency loss but also a supervised loss from the stereo depth. Their superior results show that an additional supervised loss does help boost performance.

Kuznietsov et al.~\cite{kuznietsov2017semi} alternatively solve the scale ambiguity problem of solely monocular depth estimation by adding sparse depth as a loss term to the overall loss function. Their loss function also contains a stereo loss term as in Godard~\cite{undeepvo_li2018}. Compared to this work, our approach takes advantage of the temporal constraint between pairs of sequential images at no extra cost without the need of the stereo image pair. Our model can also take as the input the RGB images only. Results with these settings are reported in section~\ref{sec:evaluation}.  While making use of stereo images limit Yang's and Godard's works to stereo sensors only, our proposed method can work with a wide range of depth sensors.


Wang et. al~\cite{wang_differential_depth_wang2018learning} incorporate a Differentiable Direct Visual Odometry (DDVO) pose predictor as an alternative to pose-CNN such that the back-propagation signals reaching the camera pose predictor can be propagated to the depth estimator. Unlike our approach, this framework requires no learnable parameters for the pose estimation branch and also enables joint training for pose and depth estimation. However, because of the nonlinear optimization in DDVO a good initialization is required, such as using a pretrained Pose-CNN to provide pose initialization, for DDVO to converge during the training.

\subsection{Depth Completion} 
There has been recent success in generating dense depth images by exploiting sparse depth input from LiDAR sensors and high resolution RGB imagery~\cite{top_depth_completion_van2019sparse, eldesokey2018confidence, jaritz2018sparse, uhrig2017sparsity} with the KITTI Depth Completion benchmark. However, these methods rely heavily on the supplied sparse depth input and the accuracy of these measurements, making them susceptible to failure with noisy sparse input samples. They additionally often struggle to extrapolate depth values where there are no LiDAR measurements \cite{shivakumar2019dfusenet}.

The method proposed by Ma et al.~\cite{fang_net_2018self} is similar to ours. In this work, pose estimation relies upon the PnP method, which is handled independently from the depth completion pipeline. PnP relies on feature detection and correspondences, which is likely to perform poorly in low texture environments and assumes that the input depth is relatively free of noise. Alternatively, our end-to-end framework utilizes temporal constraints to formulate a measure of reprojection error, providing a training signal to the pose estimation and depth completion branches simultaneously. This reprojection error signal handles the noisy sparse depth input, while the supervision from ground truth depth provides the scale for the depth estimation.

%% file: tex/3_Problem_formulation.tex
\section{Problem Formulation}
\label{sec:problem}
\textbf{Notation: } Let $I_k:  \Omega_I \subset \mathbb{R}^2  \mapsto \mathbb{R}^3$ be the color image captured at timestep $k$, where $\Omega_I$ is the domain of the color image. The camera position and orientation at $k$ is expressed as a \textit{rigid-body transformation} $\mathbf{T}_{w \to k} \in SE(3)$ with respect to the world frame $w$. A relative transformation between two consecutive frames can be represented as 
$\mathbf{T}_{k-1 \to k}  = \mathbf{T}_{w \to k} \ast  \mathbf{T}_{w \to k-1}^{-1}$. 

Let $\mathbf{D}_k: \Omega_I  \mapsto \mathbb{R}$ be the true depth map that we want to estimate and $\mathbf{d}_k: \Omega_I  \mapsto \left \{ \mathbb{R}, \emptyset \right \} $ be the corresponding 3D projection of depth measurement on the image frame, where $\emptyset$ corresponds to pixels where depth measurement is not available or invalid. Let $\mathbf{1}_{\mathbf{d}_k \not \in \emptyset}(u,v)$ be the indicator that indicates the validity of the depth measurement at $(u, v)^T \in \Omega_I$.  A sparse and noisy depth measurement can then be represented as 
\begin{equation}
    \mathbf{d}_k(u, v) = \left \{
  \begin{aligned}
    & \mathbf{D}_k(u, v) + \epsilon(u,v),  \mbox{ if } \mathbf{1}_{\mathbf{d}_k \not \in \emptyset}(u,v) = 1  \\
    & \in \emptyset, \mbox{ otherwise }  
  \end{aligned}\right.
\end{equation}
where $\epsilon (u,v) = \mathcal{N}(0, \sigma(u,v))$ is a zero mean Gaussian noise model that we assume for depth sensors in this study. We model the noise with standard deviation proportional to the ground truth depth at a given pixel image location:  $\sigma(u,v)=\mathbf{D}_k(u, v)*f$ where $f$ controls the noise level of the sensors. In experiments with the TUM dataset in section~\ref{sec:evaluation}, we set $f=50\%$. 

Given a measurement $\mathcal{S} = \{ (I_k, \mathbf{d}_k) \hspace{2mm} | \hspace{2mm} k \in [1, N] \}$, our goal is to estimate a set $\mathcal{O} = \{ (\mathbf{\hat{T}}_{k-1 \to k}, \mathbf{\hat{D}}_k) \hspace{2mm} | \hspace{2mm} k \in [1, N]\}$ that maximizes the likelihood of the measurement.

%% file: tex/4_Framework.tex
\section{Framework}
\label{sec:framework}
We develop a deep neural network to model the rigid-body pose transformation as well as the depth map. The network consists of two branches: one CNN to learn the function that estimates the depth ($\psi_d$), and one CNN to learn the function that estimates the pose ($\theta_p$). 

This network takes as input the image sequence and corresponding sparse depth maps and outputs the transformation as well as the dense depth map. During training, the two sets of parameters are simultaneously updated by the training signal which will be detailed in this section.

\begin{figure}
    \centering
    \includegraphics[width=0.95\linewidth]{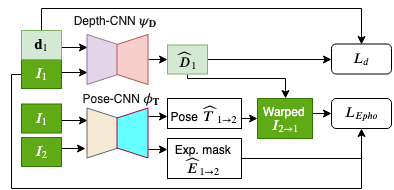}
    \caption{Network Architecture}
    \label{fig:network}
\end{figure}
The architecture is depicted in Fig.~\ref{fig:network}. We adopt the revised depth net of Ma~\cite{fang_net_2018self} as a Depth-CNN where the last layer before the ReLU is replaced by normalization and an Exponential Linear Unit function (ELU). Pose-CNN is adapted from Sfmlearner~\cite{zhou2017unsupervised}. The losses used to train the network are detailed as follows. 

 \subsection{Supervised Loss}
In this study, we assume that during the training, a semi-dense ground truth depth $\mathbf{\widetilde{D}}_k$ is known and used to supervise the network by penalizing the difference between the depth prediction $\mathbf{\hat{D}}_k = \mathbf{\psi}_{\mathbf{D}}(I)$ and itself. Note that this semi-dense ground truth depth is not necessary during the testing. The supervised loss is applied to the set of pixels with directly measured depth available. It reads:
 \begin{equation}
    \mathcal{L}_\mathbf{D} = ||\mathbf{1}_{\mathbf{\widetilde{D}}_k \neq 0}(\mathbf{\hat{D}}_k - \mathbf{\widetilde{D}}_k)||_2^2
 \end{equation}

\subsection{Photometric Loss} 
The semi-dense depth is often sparse due to the hardware and software limitations of depth sensors, making the supervised loss incapable of generating a pixel-wise cost for the estimated dense depth $\mathbf{\hat{D}}_k$. To cope with this problem, we introduce an unsupervised loss which is essentially a photometric loss that enforces the temporal constraint between two consecutive image frames. 

\textbf{Photometric Loss}

The unsupervised loss is similar to the residuals used in direct visual odometry which is often computed at different image scales. An unsupervised loss computed at scale $s$ can be represented as follows.
\begin{equation}
\mathcal{L}_{pho}^{(s)} = \sum_\mathbf{u} \sigma I(\mathbf{T}, \mathbf{u}, s)  d\mathbf{u},
    \label{eq:unsup_residual}
\end{equation}
where $\mathbf{u} = (u, v)^T \in \Omega_I$ and the residual intensity $\sigma I$ is defined by the photometric difference between pixels observed from the same 3D point between one image $I$ and its warping under the transformation $\mathbf{T}$. 



To find the corresponding transformation between the two frames to compute the loss in Eq.~\ref{eq:unsup_residual}, Ma et al.~\cite{fang_net_2018self} utilize a PnP framework which is susceptible to failure in low-texture environments. In contrast, our framework introduces another DCNN $ \mathbf{\phi}_{\mathbf{T}}(I_1, I_2)$ that estimates the pose $\mathbf{\hat{T}}_{1\to 2}$ between two consecutive frames $1$ and $2$.  In this way, both depth estimation and ego-motion are differentiable, enabling the effective training of both pose estimation and depth estimation in an end-to-end scheme. 

\textbf{Residuals}

Given the depth prediction $\mathbf{\hat{D}}_1$ of the first frame and the estimated relative pose $\mathbf{\hat{T}}_{1\to 2}$ between the current frame $I_1$ and the consecutive frame $I_2$, a warped image $I_{2\to1}$ can be generated by warping the frame $I_2$ to the previous frame $I_1$. The intensity value of a pixel $\mathbf{u}_1\in \Omega_I$  in the warped image can be computed using the pinhole camera model as follows.  
 \begin{equation}
     I_{2 \to 1}(\mathbf{u_1}) = I_2(K\mathbf{\hat{T}}_{1\to 2}\mathbf{\hat{D}}_1(\mathbf{u_1})K^{-1}\widetilde{\mathbf{u}}_1),
 \end{equation}
 where $\widetilde{\mathbf{u}}_1\in \mathbf{R}^3$ is the homogeneous representation of pixel $\mathbf{u_1}$.
 
 Similarly, the warped image $I_{1 \to 2}$ can be generated from the inverse pose $\mathbf{\hat{T}}_{1\to 2}^{-1}$:
 \begin{equation}
     I_{1 \to 2}(\mathbf{u_2}) = I_2(K\mathbf{\hat{T}}_{1\to 2}^{-1}\mathbf{\hat{D}}_2(\mathbf{u_2})K^{-1}\widetilde{\mathbf{u}}_2)
 \end{equation}
 
 The average residual is then computed from the residuals of these two warped images. The residual for scale $s$ is defined as:
 \begin{align*}
        \sigma I(\mathbf{\hat{T}}, \mathbf{u},s)= \frac{1}{2s}(&||\textbf{1}^{(s)}_{d_1==0} (I^{(s)}_{2 \to 1}(\mathbf{u}) - I^{(s)}_1(\mathbf{u}))||_1 + \\
        &||\textbf{1}^{(s)}_{d_2==0} (I^{(s)}_{1 \to 2}(\mathbf{u}) - I^{(s)}_2(\mathbf{u}))||_1)
 \end{align*}



 \subsection{Masked Photometric Loss}
The synthesis of image $I_{1\to 2}$ and $I_{2\to 1}$ implicitly assumes that the scene is static --- a condition that does not always hold. To cope with moving objects in the scene, we introduce the explainability mask network $\hat{E}$ as in~\cite{zhou2017unsupervised}, which models the confidence of each pixel in the synthesized image. A static object will have pixels with confidence of $1$. In practice, this mask network can be a DCNN branch attached to the pose network, making the whole network model trainable in an end-to-end manner. The masked photometric loss that takes into account the confidence is, therefore, capable of coping with moving objects, and formulated as
 \begin{equation}
     \mathcal{L}_{E pho} = \sum_{s}  \hat{E}^{(s)} 
     \mathcal{L}_{pho}^{(s)}
 \end{equation}

 \subsection{Smoothness Loss}
We also enforce the smoothness of the depth estimation by using a smoothness loss term which penalizes the discontinuity in the depth prediction. The $L_1$ norm of the second-order gradients is adopted as the smoothness loss $\mathcal{L}_{smo}$ as in~\cite{vijayanarasimhan2017sfm}.

\subsection{Training Loss Summarization}
In summary, the final loss for our entire framework is defined as:
\begin{equation}
    \mathcal{L}_{final} = \alpha \mathcal{L}_d + \beta \mathcal{L}_{Epho}+ \gamma \mathcal{L}_{smo} + \theta \mathcal{L}_{E}
\end{equation}
where $\alpha$, $\beta$ and $\gamma$ are weights for the corresponding losses, which are set to 1.0, 0.1, 0.1 and 0.2 empirically. An additional loss term $\mathcal{L}_{E}$ is also adopted from Zhou~\cite{zhou2017unsupervised} to avoid degeneration of the mask, where the mask can be zero to favor the minimal photometric loss.

%% file: tex/5_Experiment.tex
\section{Experimental Settings}
\subsection{Datasets}
In this study, we conduct experiments to evaluate our approach in comparison with others on two datasets. 

We first evaluate our approach on KITTI raw dataset~\cite{kitti_geiger2012we} on two different aspects: depth completion and pose estimation. We train our framework on 44,000 training data, validate on 1,000 selected data. 

For depth completion, Kitti provides a separate test set with unknown ground truth for benchmarking. The predicted depth on the test set is submitted to the Kitti's benchmark system to obtain the result given in Tab.~\ref{tab:kitti_depth_completion_rank}. 

For pose estimation, we evaluate based on three different metrics: absolute trajectory error(ATE)~\cite{sturm12iros}, relative pose error (RE)~\cite{sturm12iros} and average photometric loss. The comparison result is shown in Tab.~\ref{tab:kitti_odometry_rank}.  
We then evaluate the performance of our approach on TUM RGB-D SLAM Dataset~\cite{sturm12iros}. 
We divide this dataset into two sets: the training set consists of sequences: \textit{freiburg2 pioneer slam} and \textit{freiburg2 pioneer slam3}. The test set is \textit{freiburg2 pioneer slam2}. 

The depth completion performance is depicted in Tab.\ref{tab:tum_depth_completion_rank} and the pose estimation performance is depicted in Tab.~\ref{tab:tum_pose_rank}. 

\subsection{Implementation Details}
Our algorithm is implemented with Pytorch. We use ADAM optimizer with a momentum of 0.9 and weight decay of 0.0003. The ADAM parameters are pre-selected as $\alpha = 0.0001, \beta_1 = 0.9$ and $\beta_2 = 0.999$.  Two Tesla V100 with RAM of 32GB are used for training with the batch size of 8 and 15 epochs take around 12 hours.

For Kitti dataset, we take input images $I$ and sparse depth $d$ both with the size of $376 \times 1242$. The supervision of semi-dense depth map $D$ is formed by aggregating sparse depth from the nearby 10 frames. For TUM robot slam dataset, input images $I$ and sparse depth $d$ are both with the size of $480 \times 640$, being taken into the framework. Input images $I$ are normalized with $[\sigma, \mu]=[0.5, 0,5]$ and input sparse depth is scaled in value. Input sparse depth is scaled by a factor of 0.01 on Kitti while of 1/15 on TUM. The supervision depth $D$ is as the same size of input sparse depth while not being scaled.

\begin{figure}[ht]
    \centering
    \includegraphics[width=0.95\linewidth]{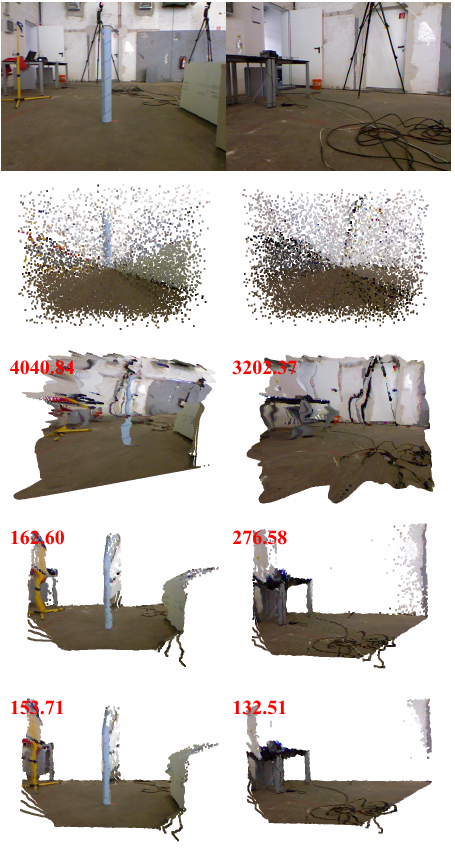}
    \caption{Depth refinement result. Numbers on depth images indicate the RMSE. Top: input images. $2^{nd}$ row: sparse, noisy input depth. $3^{rd}$ row: output depth from Sfmleaner. $4^{th}$ row: output depth from Ma~\cite{fang_net_2018self}. $5^{th}$ row: output depth from our model. Our model is more resilient to input noisy depth then Ma~\cite{fang_net_2018self}.}
    \label{fig:depth_refinement_results}
\end{figure}

%% file: tex/6_Evalutation.tex
\section{Evaluation}
\label{sec:evaluation}

 \subsection{Depth Completion}
\begin{figure*}[ht]
  \includegraphics[width=\textwidth]{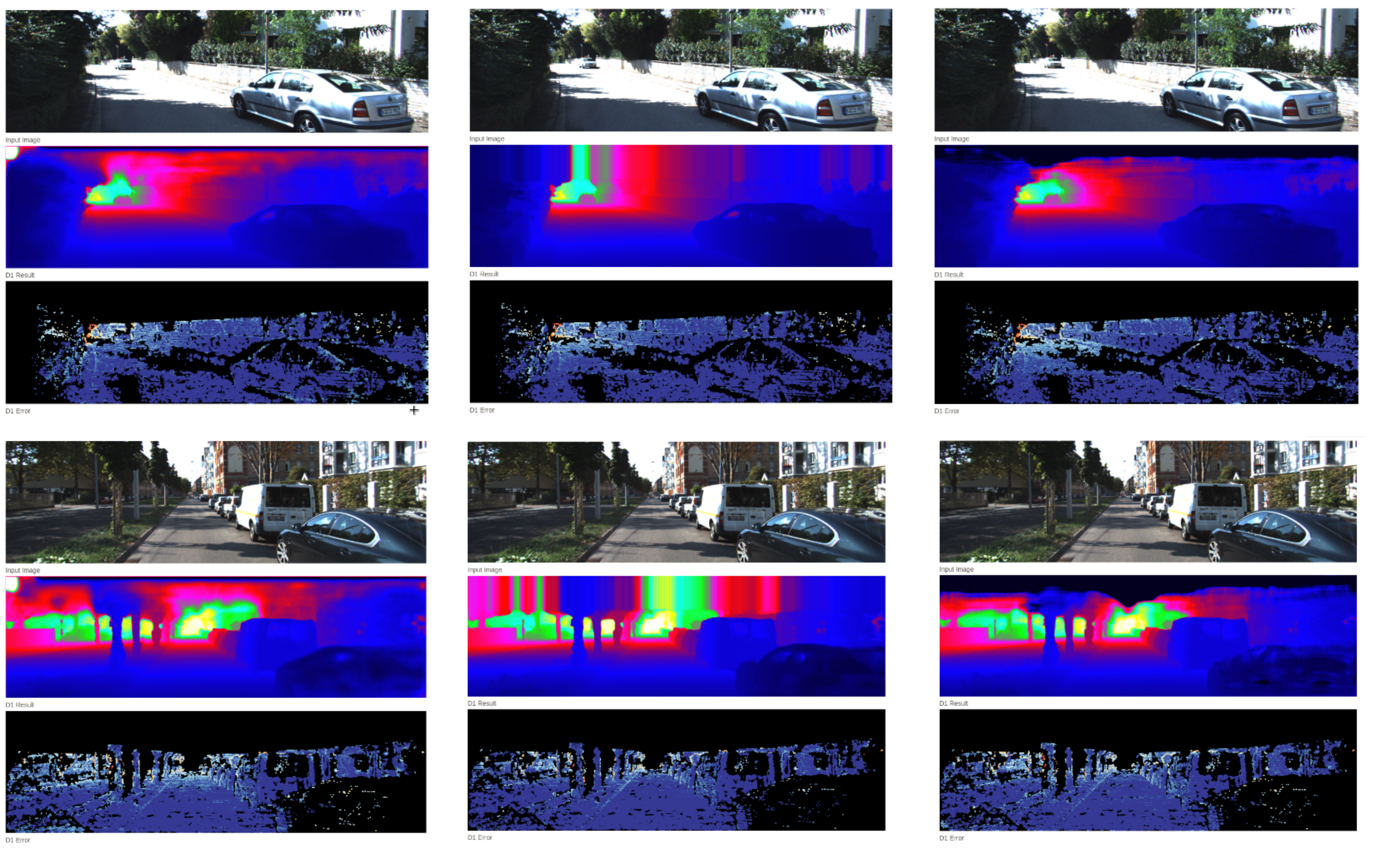}
  \caption{Qualitative results of our method (left), RGB guide\&certainty~\cite{vangansbeke2019} (middle) ranking 1st and Ma~\cite{fang_net_2018self} (right) ranking 7th.}
  \label{fig:more_depth}
\end{figure*}

 In this section, we benchmark our proposed method against other state-of-the-art methods which mostly focus on depth completion task, specifically on the KITTI~\cite{kitti_geiger2012we} Depth Completion benchmark. In Tab.~\ref{tab:kitti_depth_completion_rank}, our method achieves competitive depth completion performance and ranks only 17th in terms of RMSE error but is able to fill depth for more pixels without depth ground truth as shown in Fig.~\ref{fig:more_depth}. This is because the incorporation of both supervised loss and unsupervised loss. Moreover, the difference between our RMSE and that of the number one approach is only equivalent to $0.2\%$ of the maxinum distance on the Kitti dataset. 
 
 We focus on comparison with Ma~\cite{fang_net_2018self} since this method is close to our approach. In the KITTI dataset, they rank number $7^{th}$ while ours ranks number $17^{th}$. 
 
 The first three rows of Tab.~\ref{tab:tum_depth_completion_rank} depicts the performance of our approach in comparison with that of Ma~\cite{fang_net_2018self} and Sfmlearner on the depth completion problem in the TUM dataset. In this dataset, our proposed method performs better than that of Ma~\cite{fang_net_2018self}. 


\begin{table}[ht]
\centering
\begin{tabular}{|l|c|c|c|c|}
\hline
Method	   	&iRMSE  &iMAE   &RMSE  &MAE	\\	
\hline \hline
RGB guide $\&$ certainty	&2.19	&0.93	&772.87	&215.02\\
DL2		    &2.79	&1.25	&775.52	&245.28\\	
RGB guide $\&$ certainty  &2.35	&1.01	&775.62	&223.49\\	
DL1		    &2.26	&0.99	&777.90	&209.86\\	
DLNL			&3.95	&1.54	&785.98	&276.68\\	
RASP		    &2.60	&1.21	&810.62	&256.00\\	
Ma~\cite{fang_net_2018self}  &2.80	&1.21	&814.73	&249.95\\	
NConv-CNN-L2 (gd)	 &2.60	&1.03	&829.98	&233.26\\	
MSFF-Net		     &2.81	&1.18	&832.90	&247.15\\	
DDP			     &2.10	&0.85	&832.94	&203.96\\	
NConv-CNN-L1 (gd)	 &2.52	&0.92	&859.22	&207.77\\	
LateFusion		     &4.59	&2.25	&885.92	&347.61\\	
Spade-RGBsD		 &2.17	&0.95	&917.64	&234.81\\	
glob guide $\& $ certainty	&2.80	&1.07	&922.93	&249.11\\
RDSS			     &3.22	&1.44	&927.40	&312.23\\	
HMS-Net		     &2.93	&1.14	&937.48	&258.48\\
\hline
Ours(RGBD)		     &3.21	&1.39	&943.89 &304.17\\
Ours (d)            &4.48     &1.78   &1184.35&403.18 \\
\hline
\end{tabular}
\caption{KITTI depth completion ranking. Ours rank number $17^{th}$ as of the submission date. 
iRMSE, iMAE are in $1/km$. RMSE and MAE are in $mm$}
\label{tab:kitti_depth_completion_rank}
\end{table}



 \subsection{Pose Estimation}
Since most methods in the depth completion benchmark do not investigate joint pose estimation, we compare our trained pose estimator against Sfmlearner, and Ma~\cite{fang_net_2018self} - which is essentially PnP. Sfmlearner is already capable of simultaneously estimating both pose and depth estimation similar to our method. The parameters for the PnP algorithm are set according to Ma~\cite{fang_net_2018self}. Tab.~\ref{tab:kitti_odometry_rank} and \ref{tab:tum_pose_rank} shows that our method achieves lower photometric loss than all other mentioned methods on KITTI~\cite{kitti_geiger2012we} dataset while exhibiting very competitive ATE and RE values to the other methods. 
Compared with Sfmlearner's result stated in the paper~\cite{zhou2017unsupervised}, our depth prediction 
RMRSE error, $0.943$, is far superior to that of Sfmlearner, $6.565$. This explains why our approach achieves lower photometrics loss. 

To diversify the evaluation, we compare the same approaches on the TUM~\cite{sturm12iros} dataset. The TUM RGBD dataset is particularly challenging, as it contains large amounts of rotation, particularly at the beginning.  In addition, there are several blurred frames, as well as points of rapid rotation causing significant jump between frames.  Finally, the dataset was gathered using a Kinect camera, which uses a rolling shutter camera creating geometric distortion in the image. These challenges make TUM RGBD dataset ideal for evaluating the robustness of the pose estimation algorithms. 

We additionally benchmark monocular DSO \cite{engel2016dso} on the TUM dataset, as a representative of the state-of-the-art in monocular SLAM (Simultaneous Localization and Mapping).  Monocular methods are particularly sensitive to distortion from rolling shutters as well as calibration error.  These factors cause DSO to diverge early in the dataset. Thus, we omit its result here. 

As Tab.~\ref{tab:tum_pose_rank} depicts, our approach outperforms both Ma~\cite{fang_net_2018self} and Sfmlearner when considering ATE, RE and photometric loss as metrics. This result, along with the experiments in Tab.~\ref{tab:tum_depth_completion_rank} concludes that our approach outperforms Ma~\cite{fang_net_2018self} and Sfmlearner on both pose estimation and depth estimation in TUM dataset.

\begin{table}[h]
\centering
\begin{tabular}{|l|c|c|c|}
\hline
Method               & ATE[m]              & RE                 & Photometric Loss    \\
\hline
\hline
Ma~\cite{fang_net_2018self}  & 0.0105$\pm $0.0082 & 0.0011$\pm$0.0006 &  0.1052      \\
sfmlearner           & 0.0179$\pm $0.0110  &0.0018$\pm$0.0009   &  0.1843  \\
Groundtruth          & 0.$\pm $0.         &0.$\pm $0.               &  0.1921  \\
\hline
Ours (rgbd)          & 0.0170$\pm $0.0094      &0.0046$\pm$0.0031 & \textbf{0.0726} \\
\hline
\end{tabular}
\caption{KITTI odometry benchmark. Ours has larger ATE and RE than others but has the smallest photometric loss. }
\label{tab:kitti_odometry_rank}
\end{table}

\begin{table}[h]
\centering
\begin{tabular}{|l|c|c|c|}
\hline
Method               & ATE[m]              & RE                 & Photometric Loss    \\
\hline
\hline
Ma~\cite{fang_net_2018self}   & 0.0116$\pm $0.0067 & 0.0025$\pm$0.0022 &  0.0820      \\
sfmlearner           & 0.0152$\pm $0.0092  &0.0033$\pm$0.0021   &  0.1043  \\
Groundtruth          & 0.$\pm $0.         &0.$\pm $0.               &  0.0533  \\
\hline
Ours (rgbd)          & \textbf{0.0101$\pm $0.0051}      &\textbf{0.0021$\pm$0.0015} & \textbf{0.0369} \\
\hline
\end{tabular}
\caption{Pose estimation performance on TUM robot SLAM dataset. Ours has both smaller ATE, RE as well as photometric loss than others.}
\label{tab:tum_pose_rank}
\end{table}

\subsection{Depth Refinement}
One of the prominent problems in robotics vision is the noisy depth measurement as the result of lightweight and low-cost depth sensors. In this section, we simulate noisy input depth by adding Gaussian noise with standard deviation proportional to each depth value and evaluate the network's ability to handle this situation with different methods. Particularly, we evaluate on the TUM robot SLAM dataset by adding to each valid depth point a Gaussian noise with standard deviation of upto $50\%$ of the depth value. We then train and test with different settings and report the result in Tab.~\ref{tab:tum_depth_completion_rank}. 

For Sfmlearner, it does not use depth so the results are the same with and without input depth noise. For ours, we evaluate on two cases: one that trains without noisy depth and one trained with noisy depth. All are tested with noisy depth. Note that the input depth is also sparse as it is obtained by randomly sampling $7\%$ from the groundtruth depth. 
The qualitative results are shown in Fig.~\ref{fig:depth_refinement_results} where each number on the image refers to the RMSE in depth estimation error on the image given by corresponding approaches. Comparisons are shown on the last $4$ rows on Tab.~\ref{tab:tum_depth_completion_rank}. Both qualitative and quantitative results favor our approach's performance.

\begin{table}[t] \fontsize{6.9}{7.2}\selectfont 
\centering
\begin{tabular}{|l|c|c|c|c|c|c|c|}
\hline
Method          & Depth Input       & RMSE      & MAE      &iRMSE & iMAE \\
\hline
\hline
Sfmlearner      &  O/O     &      3436.53         &2839.16        &2659.01    &2577.77 \\
Ma~\cite{fang_net_2018self} &    O/O           &119.53        &60.01        &15.23    &8.97  \\
Ours (rgbd)     & O/O   &\textbf{118.34}&\textbf{59.82 }& \textbf{14.77}    &\textbf{8.88} \\

\hline
Sfmlearner      &  Noise / Noise      &      3574.98         &2943.25        &2712.34    &2645.31 \\
Ma~\cite{fang_net_2018self} &     Noise/Noise       &209.81         &122.24        & 60.75     &27.05  \\
Ours (rgbd)     & O/Noise       &779.582        &579.18         & 116.95            &83.81 \\
Ours (rgbd)     & Noise/Noise       &\textbf{180.63}         &\textbf{100.20}        &\textbf{45.54}      &\textbf{21.08}\\
\hline
\end{tabular}
\caption{Depth completion performance on TUM robot SLAM dataset with different settings on depth input.
O: sparse $\&$ no noise; Noise: sparse $\&$ noise. i.e: O/Noise: train with depth with no noise, test with noisy depth input. Our methods performs comparably to Ma~\cite{fang_net_2018self} while outperforms all others when input depth is sparse and noisy.}
\label{tab:tum_depth_completion_rank}
\end{table}






%% file: tex/7_Conclusion.tex
\section{Conclusion}
Depth and ego-motion (pose) estimation are essential for autonomous robots to understand the environment and avoid obstacles. However, obtaining a dense, accurate depth is challenging. Depth sensors equipped to small robot platforms such as stereo cameras are often prone to noisy while accurate sensors such as LiDars can only provide sparse depth measurement. In this work, we mitigate these constraints by introducing an end-to-end deep neural network to jointly estimate the camera pose and scene structure. We evaluate our proposed approach in comparison with the state-of-the-art approaches on Kitti and TUM datasets. The empirical results demonstrate the superior performance of our model under sparse and noisy depth input as well as the capability to work with multiple depth sensors. These capabilities are beneficial on various scenarios from autonomous vehicles to MAVs.

%% file: root.bbl
\begin{thebibliography}{10}
\providecommand{\url}[1]{#1}
\csname url@rmstyle\endcsname
\providecommand{\newblock}{\relax}
\providecommand{\bibinfo}[2]{#2}
\providecommand\BIBentrySTDinterwordspacing{\spaceskip=0pt\relax}
\providecommand\BIBentryALTinterwordstretchfactor{4}
\providecommand\BIBentryALTinterwordspacing{\spaceskip=\fontdimen2\font plus
\BIBentryALTinterwordstretchfactor\fontdimen3\font minus
  \fontdimen4\font\relax}
\providecommand\BIBforeignlanguage[2]{{%
\expandafter\ifx\csname l@#1\endcsname\relax
\typeout{** WARNING: IEEEtran.bst: No hyphenation pattern has been}%
\typeout{** loaded for the language `#1'. Using the pattern for}%
\typeout{** the default language instead.}%
\else
\language=\csname l@#1\endcsname
\fi
#2}}

\bibitem{kitti_geiger2012we}
A.~Geiger, P.~Lenz, and R.~Urtasun, ``Are we ready for autonomous driving? the
  kitti vision benchmark suite,'' in \emph{2012 IEEE Conference on Computer
  Vision and Pattern Recognition}.\hskip 1em plus 0.5em minus 0.4em\relax IEEE,
  2012, pp. 3354--3361.

\bibitem{sturm12iros}
J.~Sturm, N.~Engelhard, F.~Endres, W.~Burgard, and D.~Cremers, ``A benchmark
  for the evaluation of rgb-d slam systems,'' in \emph{Proc. of the
  International Conference on Intelligent Robot Systems (IROS)}, Oct. 2012.

\bibitem{fla2017}
\BIBentryALTinterwordspacing
K.~Mohta, M.~Watterson, Y.~Mulgaonkar, S.~Liu, C.~Qu, A.~Makineni, K.~Saulnier,
  K.~Sun, A.~Zhu, J.~Delmerico, and et~al., ``Fast, autonomous flight in
  gps-denied and cluttered environments,'' \emph{Journal of Field Robotics},
  vol.~35, no.~1, p. 101–120, Dec 2017. [Online]. Available:
  \url{http://dx.doi.org/10.1002/rob.21774}
\BIBentrySTDinterwordspacing

\bibitem{shivakumar2018real}
S.~S. Shivakumar, K.~Mohta, B.~Pfrommer, V.~Kumar, and C.~J. Taylor, ``Real
  time dense depth estimation by fusing stereo with sparse depth
  measurements,'' 2018.

\bibitem{engel2016dso}
\BIBentryALTinterwordspacing
J.~Engel, V.~Koltun, and D.~Cremers, ``Direct sparse odometry,'' \emph{CoRR},
  vol. abs/1607.02565, 2016. [Online]. Available:
  \url{http://arxiv.org/abs/1607.02565}
\BIBentrySTDinterwordspacing

\bibitem{wang2017dsostereo}
\BIBentryALTinterwordspacing
R.~Wang, M.~Schw{\"{o}}rer, and D.~Cremers, ``Stereo {DSO:} large-scale direct
  sparse visual odometry with stereo cameras,'' \emph{CoRR}, vol.
  abs/1708.07878, 2017. [Online]. Available:
  \url{http://arxiv.org/abs/1708.07878}
\BIBentrySTDinterwordspacing

\bibitem{wenxin2018}
W.~Liu, G.~Loianno, K.~Mohta, K.~Daniilidis, and V.~Kumar, ``Semi-dense
  visual-inertial odometry and mapping for quadrotors with swap constraints,''
  05 2018, pp. 1--6.

\bibitem{zhou2017unsupervised}
T.~Zhou, M.~Brown, N.~Snavely, and D.~G. Lowe, ``Unsupervised learning of depth
  and ego-motion from video,'' in \emph{Proceedings of the IEEE Conference on
  Computer Vision and Pattern Recognition}, 2017, pp. 1851--1858.

\bibitem{undeepvo_li2018}
R.~Li, S.~Wang, Z.~Long, and D.~Gu, ``Undeepvo: Monocular visual odometry
  through unsupervised deep learning,'' in \emph{2018 IEEE International
  Conference on Robotics and Automation (ICRA)}.\hskip 1em plus 0.5em minus
  0.4em\relax IEEE, 2018, pp. 7286--7291.

\bibitem{sfmlearner++_prasad2018}
V.~Prasad and B.~Bhowmick, ``Sfmlearner++: Learning monocular depth \&
  ego-motion using meaningful geometric constraints,'' \emph{arXiv preprint
  arXiv:1812.08370}, 2018.

\bibitem{top_depth_prediction_fu2018deep}
H.~Fu, M.~Gong, C.~Wang, K.~Batmanghelich, and D.~Tao, ``Deep ordinal
  regression network for monocular depth estimation,'' in \emph{Proceedings of
  the IEEE Conference on Computer Vision and Pattern Recognition}, 2018, pp.
  2002--2011.

\bibitem{top_depth_completion_van2019sparse}
W.~Van~Gansbeke, D.~Neven, B.~De~Brabandere, and L.~Van~Gool, ``Sparse and
  noisy lidar completion with rgb guidance and uncertainty,'' \emph{arXiv
  preprint arXiv:1902.05356}, 2019.

\bibitem{fang_net_2018self}
F.~Ma, G.~V. Cavalheiro, and S.~Karaman, ``Self-supervised sparse-to-dense:
  Self-supervised depth completion from lidar and monocular camera,''
  \emph{arXiv preprint arXiv:1807.00275}, 2018.

\bibitem{hirschmuller2008stereo}
H.~Hirschmuller, ``Stereo processing by semiglobal matching and mutual
  information,'' \emph{IEEE Transactions on pattern analysis and machine
  intelligence}, vol.~30, no.~2, pp. 328--341, 2008.

\bibitem{einecke2015multi}
N.~Einecke and J.~Eggert, ``A multi-block-matching approach for stereo,'' in
  \emph{2015 IEEE Intelligent Vehicles Symposium (IV)}.\hskip 1em plus 0.5em
  minus 0.4em\relax IEEE, 2015, pp. 585--592.

\bibitem{scharstein2002taxonomy}
D.~Scharstein and R.~Szeliski, ``A taxonomy and evaluation of dense two-frame
  stereo correspondence algorithms,'' \emph{International journal of computer
  vision}, vol.~47, no. 1-3, pp. 7--42, 2002.

\bibitem{torr1999feature}
P.~H. Torr and A.~Zisserman, ``Feature based methods for structure and motion
  estimation,'' in \emph{International workshop on vision algorithms}.\hskip
  1em plus 0.5em minus 0.4em\relax Springer, 1999, pp. 278--294.

\bibitem{nister2004visual}
D.~Nist{\'e}r, O.~Naroditsky, and J.~Bergen, ``Visual odometry,'' in
  \emph{Proceedings of the 2004 IEEE Computer Society Conference on Computer
  Vision and Pattern Recognition, 2004. CVPR 2004.}, vol.~1.\hskip 1em plus
  0.5em minus 0.4em\relax Ieee, 2004, pp. I--I.

\bibitem{badino2013visual}
H.~Badino, A.~Yamamoto, and T.~Kanade, ``Visual odometry by multi-frame feature
  integration,'' in \emph{Proceedings of the IEEE International Conference on
  Computer Vision Workshops}, 2013, pp. 222--229.

\bibitem{lucas1981iterative}
B.~D. Lucas, T.~Kanade, \emph{et~al.}, ``An iterative image registration
  technique with an application to stereo vision,'' 1981.

\bibitem{irani1999direct}
M.~Irani and P.~Anandan, ``About direct methods,'' in \emph{International
  Workshop on Vision Algorithms}.\hskip 1em plus 0.5em minus 0.4em\relax
  Springer, 1999, pp. 267--277.

\bibitem{alismail2017direct}
H.~Alismail, M.~Kaess, B.~Browning, and S.~Lucey, ``Direct visual odometry in
  low light using binary descriptors,'' \emph{IEEE Robotics and Automation
  Letters}, vol.~2, no.~2, pp. 444--451, 2017.

\bibitem{cao2018estimating}
Y.~Cao, Z.~Wu, and C.~Shen, ``Estimating depth from monocular images as
  classification using deep fully convolutional residual networks,'' \emph{IEEE
  Transactions on Circuits and Systems for Video Technology}, vol.~28, no.~11,
  pp. 3174--3182, 2018.

\bibitem{fu2017compromise}
H.~Fu, M.~Gong, C.~Wang, and D.~Tao, ``A compromise principle in deep monocular
  depth estimation,'' \emph{arXiv preprint arXiv:1708.08267}, 2017.

\bibitem{undeepVO_godard2017unsupervised}
C.~Godard, O.~Mac~Aodha, and G.~J. Brostow, ``Unsupervised monocular depth
  estimation with left-right consistency,'' in \emph{Proceedings of the IEEE
  Conference on Computer Vision and Pattern Recognition}, 2017, pp. 270--279.

\bibitem{yang2018deep}
N.~Yang, R.~Wang, J.~Stuckler, and D.~Cremers, ``Deep virtual stereo odometry:
  Leveraging deep depth prediction for monocular direct sparse odometry,'' in
  \emph{Proceedings of the European Conference on Computer Vision (ECCV)},
  2018, pp. 817--833.

\bibitem{kuznietsov2017semi}
Y.~Kuznietsov, J.~Stuckler, and B.~Leibe, ``Semi-supervised deep learning for
  monocular depth map prediction,'' in \emph{Proceedings of the IEEE Conference
  on Computer Vision and Pattern Recognition}, 2017, pp. 6647--6655.

\bibitem{wang_differential_depth_wang2018learning}
C.~Wang, J.~Miguel~Buenaposada, R.~Zhu, and S.~Lucey, ``Learning depth from
  monocular videos using direct methods,'' in \emph{Proceedings of the IEEE
  Conference on Computer Vision and Pattern Recognition}, 2018, pp. 2022--2030.

\bibitem{eldesokey2018confidence}
A.~Eldesokey, M.~Felsberg, and F.~S. Khan, ``Confidence propagation through
  cnns for guided sparse depth regression,'' \emph{arXiv preprint
  arXiv:1811.01791}, 2018.

\bibitem{jaritz2018sparse}
M.~Jaritz, R.~De~Charette, E.~Wirbel, X.~Perrotton, and F.~Nashashibi, ``Sparse
  and dense data with cnns: Depth completion and semantic segmentation,'' in
  \emph{2018 International Conference on 3D Vision (3DV)}.\hskip 1em plus 0.5em
  minus 0.4em\relax IEEE, 2018, pp. 52--60.

\bibitem{uhrig2017sparsity}
J.~Uhrig, N.~Schneider, L.~Schneider, U.~Franke, T.~Brox, and A.~Geiger,
  ``Sparsity invariant cnns,'' in \emph{2017 International Conference on 3D
  Vision (3DV)}.\hskip 1em plus 0.5em minus 0.4em\relax IEEE, 2017, pp. 11--20.

\bibitem{shivakumar2019dfusenet}
S.~S. Shivakumar, T.~Nguyen, S.~W. Chen, and C.~J. Taylor, ``Dfusenet: Deep
  fusion of rgb and sparse depth information for image guided dense depth
  completion,'' 2019.

\bibitem{vijayanarasimhan2017sfm}
S.~Vijayanarasimhan, S.~Ricco, C.~Schmid, R.~Sukthankar, and K.~Fragkiadaki,
  ``Sfm-net: Learning of structure and motion from video,'' \emph{arXiv
  preprint arXiv:1704.07804}, 2017.

\bibitem{vangansbeke2019}
W.~Van~Gansbeke, D.~Neven, B.~De~Brabandere, and L.~Van~Gool, ``Sparse and
  noisy lidar completion with rgb guidance and uncertainty,'' \emph{arXiv
  preprint arXiv:1902.05356}, 2019.

\end{thebibliography}
